\documentclass[conference]{IEEEtran}
\IEEEoverridecommandlockouts

\usepackage{cite}
\usepackage{amsmath, amssymb, amsfonts}
\usepackage{graphicx}
\usepackage{textcomp}
\usepackage{xcolor}
\usepackage{booktabs}
\usepackage{url}
\usepackage{multirow}
\usepackage{hyperref}
\usepackage{xurl}
\usepackage{tcolorbox}
\usepackage{mathptmx}
\usepackage{newtxtext}

\newcommand{\dpos}[1]{\textcolor{green!50!black}{$+#1$}}
\newcommand{\dneg}[1]{\textcolor{red!75!black}{$-#1$}}

\def\BibTeX{{\rm B\kern-.05em{\sc i\kern-.025em b}\kern-.08em T\kern-.1667em\lower.7ex\hbox{E}\kern-.125emX}}

\begin{document}
\title{Statutory Article Retrieval from Arabic Judicial Narratives: Family Law and Law of Evidence in Saudi Arabia}

\author{\IEEEauthorblockN{Zarah Shibli}
\IEEEauthorblockA{\textit{Department of Computer Science} \\
\textit{King Abdulaziz University}\\
Jeddah, Saudi Arabia \\
zshibli0002@stu.kau.edu.sa}
\and
\IEEEauthorblockN{Areej Alhothali}
\IEEEauthorblockA{\textit{Department of Computer Science} \\
\textit{King Abdulaziz University}\\
Jeddah, Saudi Arabia \\
aalhothali@kau.edu.sa}
\and
\IEEEauthorblockN{Lulwah Alharigy}
\IEEEauthorblockA{\textit{Department of Computer Science} \\
\textit{King Abdulaziz University}\\
Jeddah, Saudi Arabia \\
lalharigy@kau.edu.sa}

}

\maketitle

\begin{abstract}
Reliable legal information retrieval must ground legal reasoning in authoritative statutory sources. In Arabic judicial settings, long factual narratives must be mapped to concise articles that express abstract legal rules, creating a \textit{narrative-to-legal-rule abstraction gap}. Using a benchmark of 200 synthetic Saudi Arabic judicial cases and 381 articles from the Family Law and the Law of Evidence, we compare word-level and char-4 BM25, zero-shot dense encoders, per-law retrieval, dense-sparse reciprocal rank fusion (RRF), and a proposed Decompose-and-Refine (DaR) pipeline. In DaR, a large language model splits each narrative into disputed legal issues and generates, for each issue, three retrieval channels: a statute-style paraphrase, legal keywords, and a candidate taxonomy chapter. On the whole corpus, the instructed Qwen3-Embedding-8B dense encoder reaches an MRR@10 of 0.849, the best result of any non-decomposed system. Under per-law scoring, this advantage reflects Family Law only: restricted to Evidence articles, the same encoder drops to an MRR@10 of 0.087, and fusion with BM25 raises it only to 0.207. DaR changes this picture, raising Evidence MRR@10 to 0.696 and Family MRR@10 to 0.864. Without decomposition, the two laws favor different strategies; under DaR, the  same retriever achieves the best MRR@10 on both.

\end{abstract}


\section{Introduction}
\label{sec:introduction}

Litigants, legal aid workers, and judges must identify the statutory provisions that govern a set of facts before a dispute can be adjudicated. The recent promulgation of Saudi Arabia's Family Law (Personal Status)~\cite{saudi2022familylaw} and Law of Evidence~\cite{saudi2021evidencelaw} has raised the stakes of reliable legal information retrieval, particularly where lengthy judicial narratives must be connected to the governing provisions. Large language models (LLMs) can hallucinate plausible but unsupported statements, which limits their standalone use in high-stakes legal settings~\cite{mohamed2026arabljp}. Retrieval-augmented generation (RAG) mitigates this by grounding generation in retrieved text~\cite{lewis2020rag}, but its value depends directly on retrieval quality. Recent Saudi-focused RAG systems emphasize end-to-end generation over small query sets~\cite{elkafrawy2026lawsuit}, which leaves article-level retrieval behavior underexamined.

A central obstacle is the mismatch between judicial narratives, which are full of procedural history and legally irrelevant detail, and statutory articles, which state concise rules, conditions, and consequences. We call this the \textit{narrative-to-legal-rule abstraction gap}. It is related to the broader challenge of matching natural-language descriptions to legal provisions~\cite{louis2022statutory}. Sparse, dense, hybrid, and iterative retrieval-generation methods have been explored in legal and Arabic settings~\cite{alsubhi2025optimizing, hrimech2025retrieval,  aboasal2026morphological, shao2023enhancing}. To our knowledge, their effectiveness has not been tested on long, formal Saudi Arabic narratives targeting individual articles, and it is unknown whether one configuration serves both statutory domains.

We address these gaps with a controlled article-level evaluation over 200 narratives and 381 articles, guided by three research questions. \textbf{RQ1:} How do sparse (word and char-4 BM25), zero-shot dense, and hybrid (dense-sparse RRF) methods compare on long Arabic judicial narratives? \textbf{RQ2:} Do the Family Law and the Law of Evidence require different retrieval strategies? \textbf{RQ3:} Does the DaR (Decompose-and-Refine) pipeline improve over the strongest non-decomposed baseline for each law?

Our contributions are (i)~an article-level evaluation of sparse, dense, and hybrid retrieval, including sixteen zero-shot encoders, on Saudi Arabic narratives; (ii)~evidence that the two laws behave differently under the same retrievers; and (iii)~an adaptation of Decompose-and-Refine to long narratives.

\section{Related Work}
\label{sec:related_work}

\textbf{Arabic and Saudi legal NLP.} Arabic is morphologically rich and comparatively under-resourced~\cite{antoun2020arabert}. Work on Saudi law has concentrated on classification, summarization, and judgment prediction: legal-reasoning benchmarks~\cite{hijazi2024arablegaleval}, a Board of Grievances corpus~\cite{sayed2026bridging}, commercial-court cases with regulation identification~\cite{abushairah2025alarb}, judgment prediction~\cite{almalki2025saudi,mohamed2026arabljp}, and the Lawsuit AraRAG pipeline~\cite{elkafrawy2026lawsuit}. To our knowledge, none of these studies targets article-level retrieval for Family Law or divorce-establishment (\textit{Ithbat Talaq}) narratives.

\textbf{Statutory article retrieval.} The Belgian Statutory Article Retrieval Dataset (BSARD)~\cite{louis2022statutory} introduced statutory article retrieval on French legal questions and showed that substantial room for improvement remains for both lexical and dense retrievers. Later studies address Arabic legal settings, including RAG over the Moroccan family code~\cite{hrimech2025retrieval}, Arabic legal retrieval with morphological segmentation~\cite{aboasal2026morphological}, and Arabic RAG pipelines more broadly~\cite{alsubhi2025optimizing,alrasheed2025evaluating}. 
These studies use short queries within one topical domain, whereas Saudi divorce narratives are long (265 words on average) and may cite two laws. To our knowledge, the effectiveness of statutory retrieval for long, cross-law judicial narratives therefore remains unexplored (RQ1).

\textbf{Sparse, dense, and hybrid retrieval.} BM25~\cite{robertson2009probabilistic} is a strong sparse baseline, dense bi-encoders~\cite{karpukhin2020dense} add semantic generalization, and instruction-conditioned encoders~\cite{su2023one} allow a single model to serve multiple retrieval tasks by prepending a natural language task instruction to the query, so that no separate model or fine-tuning is needed per task. RRF~\cite{cormack2009reciprocal} combines rankings without training, and gains from BM25--dense fusion have been reported for Arabic and legal text~\cite{aboasal2026morphological,alsubhi2025optimizing}. The gain depends on the relative strength of the fused retrievers, so it is unknown whether fusion helps the strongest dense retriever on long Arabic narratives, or whether per-law routing is needed (RQ1 and RQ2).

\textbf{Query expansion, decomposition, and reranking.} HyDE~\cite{gao2023hyde} and Query2doc~\cite{wang2023query2doc} expand a query with an LLM-generated pseudo-document, decomposition methods split a complex question into sub-questions~\cite{khot2023decomposed}, iterative retrieval--generation~\cite{shao2023enhancing} alternates between retrieving and generating, and RankRAG~\cite{yue2024rankrag} reranks a candidate pool. These methods target short, focused questions, whereas a long narrative may raise several distinct issues, and its target is a short rule reached by legal inference. Closest to our work, concurrent research proposes Decompose-and-Refine for Korean multi-hop legal question answering~\cite{lee2026decompose}: it splits a question into sub-questions and generates one statute-aligned parametric query per sub-question to select a single provision. We adapt it to a different input: we decompose a long narrative into disputed issues and use three parallel per-issue channels (paraphrastic, lexical, and taxonomy-prior) fused by RRF for article-level retrieval on Arabic statutory text (RQ3).

\section{Materials and Methods}
\label{sec:materials_methods}

\subsection{Dataset and Corpus}
\label{subsec:dataset}

The benchmark is taken unmodified from a synthetic Saudi family-court simulation corpus \cite{shibli2026synthetic}; we do not regenerate or alter the underlying cases or citations. The dataset contains 200 Arabic judicial narratives covering divorce-establishment (\textit{Ithbat Talaq}) scenarios of the explicit, implicit, and conditional types (Table~\ref{tab:dataset-stats}), validated in the source corpus through legal-expert review. Each case has a procedural \textit{Preface}, a \textit{Fact}, a \textit{Reasoning}, and a \textit{Judgment} field; only \textit{Fact} is used as the query, excluding the formulaic \textit{Preface}. By construction in the source corpus, citations are extracted from \textit{Reasoning} and \textit{Judgment} while the query uses only Fact, keeping query and label text disjoint.

\begin{equation}
    R_{i}^{\mathrm{cite}}= \operatorname{ExtractArticles}(\operatorname{Reasoning}
    _{i}, \operatorname{Judgment}_{i}).
\end{equation}

Table~\ref{tab:dataset-stats} summarizes the corpus. The task is difficult because of the abstraction gap between long narratives and short, single-topic articles: articles average about 41 words, whereas \textit{Fact} fields average 265 words (about 6.5:1), so a retriever must compress substantial procedural and factual detail into the concise rule to which it maps.

\begin{table}[htb]
    \centering
    \caption{Dataset and corpus statistics.}
    \label{tab:dataset-stats} 
    \scriptsize
    \begin{tabular}{lr}
        \toprule \textbf{Statistic}                                                             & \textbf{Value}  \\
        \midrule \multicolumn{2}{l}{\textit{\textbf{Cases (queries)}}}                                    \\
        \quad Total cases                                                                       & 200             \\
        \quad \emph{Sarih} (explicit) / \emph{kinaya} (implicit) / \emph{muallaq} (conditional) & 66 / 66 / 68    \\
        \quad Mean reference-set size (range)                                                   & $\sim$5 (1--12) \\
        \quad Cases with $\geq$1 Family citation                                                & 200 (100.0\%)   \\
        \quad Cases with $\geq$1 Evidence citation                                              & 161 (80.5\%)    \\
        \quad Mean \textit{Fact} length (words)                                                 & 265             \\
        \midrule \multicolumn{2}{l}{\textit{\textbf{Statutory corpus (articles)}}}                        \\
        \quad Total articles                                                                    & 381             \\
        \quad Family Law articles                                                               & 252             \\
        \quad Law of Evidence articles                                                          & 129             \\
        \quad Mean article length (words)                                                       & $\sim$41        \\
        \bottomrule
    \end{tabular}
\end{table}

\subsection{Problem Formulation}
\label{subsec:problem_formulation}
Let $\mathcal{A}=\{a_{1},\ldots,a_{N}\}$, with $N=381$, be the law corpus, and let  $\mathcal{Q}=\{q_{1},\ldots,q_{M}\}$, with $M=200$, be the case queries, where  $q_{i}=\operatorname{Fact}_{i}$ for $i\in\{1,\ldots,M\}$. For a retrieval cutoff $K$, 
a configuration returns the ranked list  $\hat{R}_{i}^{K}=[\hat{a}_{i,1},\ldots,\hat{a}_{i,K}]$ for query $i$, where  $k\in\{1,\ldots,K\}$ indexes the rank position in the returned list, so  $\hat{a}_{i,k}$ is the $k$-th retrieved article for query $i$. This list is evaluated  against $R_{i}^{\mathrm{cite}}$, the set of ground-truth articles cited by case $i$. Because cases cite multiple articles, retrieval is treated as a multi-label task.

Each article is one retrieval unit; no chunking is applied, so conditions, exceptions, and consequences stay together. The dense article representation prepends hierarchical context: \texttt{law\_name - chapter\_title: article\_body}.

\subsection{Retrieval Configurations}
\label{subsec:retrieval_configs}

The experiments are organized into a sequence of retrieval configurations, beginning with lexical and dense baselines and progressively introducing per-law restriction, dense--sparse rank fusion, and the Decompose-and-Refine pipeline.

\textbf{E1 (Lexical search (sparse)).} BM25 on the full \textit{Fact} field after Arabic normalization (diacritic and tatweel removal, character-variant standardization, whitespace collapse, tokens shorter than two characters dropped). Word-level tokenization with light affix stripping is compared against char-4.

\textbf{E2 (Semantic Search (Dense)).} Each Fact field is encoded as a single dense query; query and article embeddings are $L_{2}$-normalized and scored by exact cosine similarity. No model is fine-tuned; all sixteen encoders are evaluated zero-shot.

\textbf{E3 (Retrieval enhancement).} Two enhancements are examined: per-law restriction (E3a) and reciprocal rank fusion (E3b).

\textbf{E3a (Per-law restriction).} Retrieval is restricted to one law's articles at a time (252 Family or 129 Evidence), so a system's ranking within a law is not diluted by competing articles from the other codification. This is the setting the DaR routes (E4) operate in, and the fair-comparison baseline against which DaR is measured.

\medskip

\textbf{E3b (Dense--sparse RRF).} Rankings are combined by reciprocal rank fusion~\cite{cormack2009reciprocal}:
\begin{equation}
    S_{\mathrm{RRF}}(a) = \sum_{r\in\mathcal{R}}
    \frac{1}{k+\operatorname{rank}_{r}(a)}
    \label{eq:rrf}
\end{equation}
where $\mathcal{R}$ is the set of rankers and $\operatorname{rank}_{r}(a)$ is the rank of article $a$ under ranker $r$. We set $k=60$, the default of the original RRF paper. Four whole-corpus combinations are tested: lexical-only (word $+$ char-4 BM25), two dense--sparse pairs (word BM25 $+$ instructed 8B; word BM25 $+$ instructed 4B), and a four-way fusion (word $+$ char-4 $+$ 8B $+$ 4B). RRF is a training-free method to test whether lexical and dense signals are complementary.

\begin{figure}[t]
    \centering
    \includegraphics[width=1\linewidth]{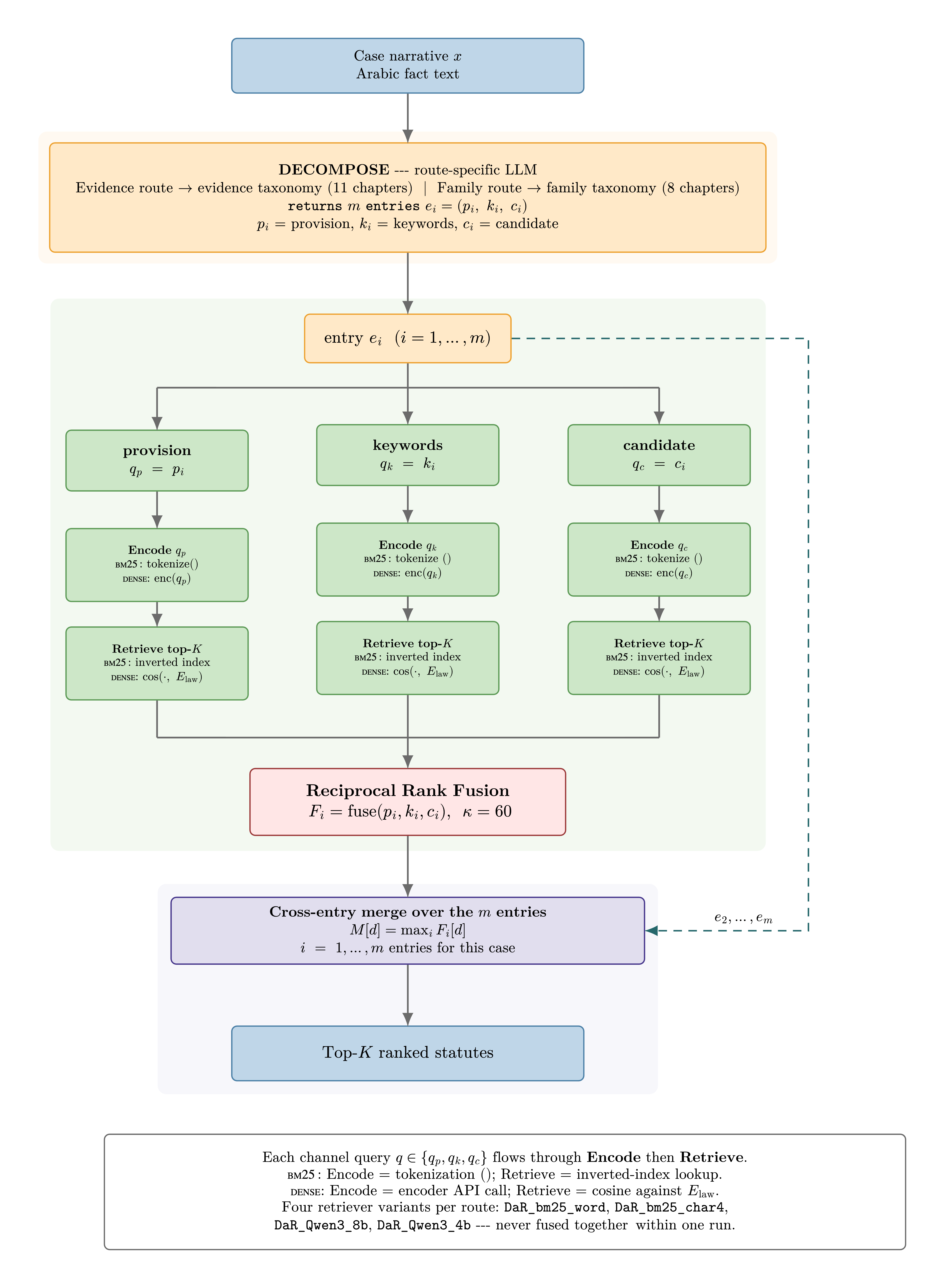}
    \caption{The Decompose-and-Refine (DaR) pipeline: an LLM decomposes a case narrative into $m$ issue entries, each retrieved through three channels and fused by RRF, then merged across issues by a per-article max score.}
    \label{fig:experimental-configurations}
\end{figure}

\textbf{E4 (Decompose-and-Refine, DaR).} We adapt the Decompose-and-Refine paradigm of Lee et al.~\cite{lee2026decompose}, originally designed for single-provision parametric retrieval in Korean multi-hop legal question answering, to article-level retrieval from long narratives. A single narrative can raise several legally distinct disputes, and the governing articles must be reached by legal inference rather than by direct passage matching (Section~\ref{sec:related_work}).  E4 runs a single constrained LLM call per case per  law route. The model reads the \textit{Fact} field, identifies the legally distinct disputes the narrative raises under that law, and returns one JSON entry per dispute, each carrying a disputed element and three retrieval channels; the entries are then retrieved and fused as described below (Fig.~\ref{fig:experimental-configurations}). The same facts are decomposed independently under each route, so a case yields one family decomposition and one evidence decomposition.

\emph{Decomposition.} A constrained LLM reads the \textit{Fact} field of a case and returns between one and five entries, one per disputed issue, in a single JSON object. Each entry contains a \emph{disputed element} (a short description of the issue) and three retrieval channels:
\begin{itemize}
    \item \emph{provision}: a statute-style paraphrase of the rule that
    governs the issue;
    \item \emph{keywords}: 5--10 requested legal terms;
    \item \emph{candidate}: a guessed location in the statute, written as
    \texttt{\textless{}chapter\textgreater{} :: \textless{}issue
    label\textgreater{}}, where the chapter is chosen from a fixed taxonomy.
\end{itemize}
The other fields of an entry are not used for retrieval. The taxonomy has 11 chapters for the Law of Evidence and 8 for the Family Law.

The prompt template is shown in Fig.~\ref{fig:dar-prompt}. One template is instantiated per law route: the bracketed slots are filled with the route's persona (Saudi judge specializing in the Law of Evidence or the Family Law), the corresponding issue type (evidentiary or substantive legal), and that route's chapter taxonomy, yielding two distinct prompts. The model is asked only to produce query formulations, not to retrieve or rank articles, and it returns a single JSON object with no surrounding prose. The schema is fixed: each entry carries a \emph{disputed element}, a \emph{provision}, \emph{keywords}, and a \emph{candidate}, with an additional \emph{family issue type} field on the family route only.

\begin{figure}[ht]
\begin{tcolorbox}
    [colback=gray!5,
    colframe=black!80, fonttitle=\small\bfseries]     
    \footnotesize
    \noindent
    You are a Saudi judge specializing in \texttt{[Law of Evidence / Family Law]}.
    Read the case facts and identify the disputed \texttt{[evidentiary / substantive
    legal]} issues the case raises. For each issue, produce from 1 to 5 structured
    query formulations used to retrieve the applicable statutory article.

    \vspace{0.4em}
    \noindent
    \textbf{Available taxonomy (candidate channel only):} \\ (\texttt{[11-chapter
    Evidence taxonomy / 8-chapter Family taxonomy]}).

    \vspace{0.3em}
    \noindent
    \textbf{Return JSON only:}\\ \texttt{\{"entries":[\{"disputed\_element":"...",
    "parametric\_provision":"...", \underline{[family route only: "family\_issue\_type":"...",]}
    "parametric\_sparse\_keywords":["...","..."], "parametric\_statute\_candidate":"\textless{}location\textgreater{}
    :: \textless{}legal\_issue\textgreater{}"\}]\}}
\end{tcolorbox}
\caption{Decompose-and-Refine prompt template (both routes).}
\label{fig:dar-prompt}
\end{figure}

\emph{Retrieval and fusion.} Each channel is used as an independent query against the corpus of the corresponding law. For each issue, the three channel rankings are fused using Reciprocal Rank Fusion (RRF; Eq.~\eqref{eq:rrf}) with $k{=}60$. Rankings across different issues are then merged into a single case-level ranking by assigning each article its highest fusion score across all issues in the case. Consequently, an article remains in contention at the case level if it is retrieved as relevant to at least one issue. The resulting ranking is truncated at $K$. The channel queries are retrieved with one of four retrievers: word BM25, char-4 BM25, and the  instructed Qwen3-Embedding-4B and -8B encoders, giving four DaR systems per route.

\emph{Implementation.} The decomposition LLM is \texttt{google/gemini-3.6-flash}~\cite{google2026gemini36flash} at temperature 0. We chose it for its speed and cost efficiency at the scale of our experiments, since decomposition requires two LLM calls per case, one per route. In a pilot comparison, it recovered a similar set of retrieval-relevant issues to larger models while providing substantially lower latency.

\subsection{Embedding Models and Evaluation}
\label{subsec:models_metrics}

Sixteen zero-shot dense encoders spanning four paradigms are evaluated on E2: Arabic-specific encoders (AraBERT-v2, AraBERT-Large-v2 \cite{antoun2020arabert}, GATE-AraBERT \cite{nacar2025gategeneralarabictext}, Arabic-Triplet-Matryoshka-V2 \cite{nacar2025gategeneralarabictext}), multilingual contrastive encoders (multilingual-E5 \cite{wang2024multilinguale5textembeddings}, BGE-M3 \cite{chen2024bge}, Snowflake-Arctic-Embed-L-v2.0 \cite{yu2024arcticembed}), proprietary API embeddings (Cohere-Embed-Multilingual-v3.0\cite{cohere2023embedv3}, Cohere-Embed-v4.0\cite{cohere2025embedv4}, Gemini-Embedding-2\cite{google2026geminiembedding2}, OpenAI text-embedding-3-small/large), and decoder-based embeddings (Qwen3-Embedding-4B/8B \cite{zhang2025qwen3embedding}, each instructed and non-instructed). Only the two instructed Qwen3 encoders, the strongest encoders on E2, are carried into the DaR route (E4). For the instructed variants, every query is prefixed with a fixed Arabic instruction conditioning the encoder toward the retrieval intent (asking it to represent the query so as to reflect the legal issue and applicable statutory rules, targeting the Family Law or Law of Evidence); non-instructed variants receive the raw \textit{Fact} text with no prefix. Because both variants are evaluated at each of the two model sizes, comparing instructed against non-instructed models within a fixed size (Section~\ref{subsec:res_e1e2}) isolates the effect of the instruction prefix from that of model scale. E1--E4 are evaluated at $K\in\{1,3,5,7,10,20,50\}$ using Precision, Recall, F1, Hit Rate, and MRR, macro-averaged over cases. All metrics are computed separately for the Family Law and the Law of Evidence rather than over the pooled 381-article corpus. For whole-corpus configurations (E1, E2, and E3b), the system first produces a single ranking over all 381 articles; for evaluation under a given law, this ranking is then filtered to retain only articles from that law before the metrics are computed. In contrast, law-restricted configurations (E3a and E4) retrieve exclusively from the corpus of the law under evaluation. For each law, only cases whose reference set contains at least one article from that law are included in the macro-average. All 200 cases contain at least one Family Law article, while only 161 contain at least one Law of Evidence article; accordingly, Family metrics are averaged over 200 cases and Evidence metrics over 161. The same filtering process is applied to every system to ensure that systems compared within each law are evaluated on identical case sets.

\section{Results and Discussion}
\label{sec:results}

We use MRR@10 as the main ranking metric and report R@10 (and R@20 where useful) alongside it. Family metrics are averaged over 200 cases and Evidence metrics over the 161 cases that cite an Evidence article. Table~\ref{tab:summary} summarizes the best system of each method family under per-law scoring; the remaining tables give the full results. Three findings emerge: whole-corpus scores hide a failure on Evidence; without decomposition, no method retrieves Evidence articles well; and DaR is the only configuration that does.

\subsection{E1--E2: Sparse and Dense Retrieval (RQ1)}
\label{subsec:res_e1e2}

Table~\ref{tab:e2} reports the sparse baselines and the sixteen zero-shot encoders on the whole 381-article corpus. Word-level BM25 outperforms char-4 BM25 on MRR@10 ($0.361$ vs.\ $0.134$), whereas char-4 has higher R@20. The best encoder, instructed Qwen3-Embedding-8B, reaches MRR@10 $0.849$, and the instruction prefix improves both model sizes ($0.803\to0.849$ for 8B; $0.762\to0.836$ for 4B). The best proprietary encoder, Gemini-Embedding-2, reaches $0.624$. Both AraBERT variants are nearly unusable, which is expected because they are not trained as sentence encoders.

The whole-corpus advantage is a Family effect. Across the sixteen encoders the Family MRR@10 column matches the whole-corpus column within $0.005$, while Evidence R@10 is exactly $0.000$ for nine encoders and at most $0.019$ for any encoder. For instructed 8B, Evidence MRR@10 is $0.001$ against $0.849$ on Family. Whole-corpus metrics are therefore not a usable diagnostic for Evidence, and we score each law separately from here on.

\begin{table*}[tb]
    \centering
    \caption{E1--E2 whole-corpus retrieval: sparse baselines and sixteen
    zero-shot encoders. $n{=}200$ (whole corpus, Family), $n{=}161$ (Evidence).
    Bold: best per column.}
    \label{tab:e2}
    \scriptsize
    \setlength{\tabcolsep}{3.5pt}
    \begin{tabular}{lccc|cc|cc}
        \toprule
        & \multicolumn{3}{c|}{\textbf{Whole corpus}} & \multicolumn{2}{c|}{\textbf{Family}} & \multicolumn{2}{c}{\textbf{Evidence}} \\
        \cmidrule(lr){2-4}\cmidrule(lr){5-6}\cmidrule(lr){7-8}
        \textbf{Model} & \textbf{R@10} & \textbf{R@20} & \textbf{MRR@10} & \textbf{R@10} & \textbf{MRR@10} & \textbf{R@10} & \textbf{MRR@10} \\
        \midrule
        \multicolumn{8}{l}{\textit{\textbf{Sparse (BM25)}}} \\
        BM25 (word)  & 0.258 & 0.379 & 0.361 & 0.351 & 0.352 & 0.049 & 0.025 \\
        BM25 (char-4) & 0.202 & 0.434 & 0.134 & 0.249 & 0.110 & \textbf{0.082} & \textbf{0.033} \\
        \midrule
        \multicolumn{8}{l}{\textit{\textbf{Arabic-specific encoders}}} \\
        AraBERT-v2 & 0.003 & 0.003 & 0.004 & 0.001 & 0.001 & 0.004 & 0.004 \\
        AraBERT-Large-v2 & 0.024 & 0.067 & 0.021 & 0.037 & 0.018 & 0.008 & 0.005 \\
        GATE-AraBERT & 0.237 & 0.393 & 0.327 & 0.339 & 0.327 & 0.000 & 0.000 \\
        Arabic-Triplet-Matryoshka-V2 & 0.380 & 0.542 & 0.443 & 0.525 & 0.443 & 0.000 & 0.000 \\
        \midrule
        \multicolumn{8}{l}{\textit{\textbf{Multilingual contrastive encoders}}} \\
        E5 (multilingual) & 0.208 & 0.313 & 0.285 & 0.276 & 0.285 & 0.000 & 0.000 \\
        BGE-M3 & 0.396 & 0.514 & 0.525 & 0.565 & 0.525 & 0.000 & 0.000 \\
        Snowflake-Arctic-Embed-L & 0.324 & 0.462 & 0.360 & 0.444 & 0.360 & 0.016 & 0.003 \\
        \midrule
        \multicolumn{8}{l}{\textit{\textbf{Proprietary API embeddings}}} \\
        Cohere-Embed-Multilingual-v3.0 & 0.341 & 0.462 & 0.401 & 0.458 & 0.401 & 0.000 & 0.000 \\
        Cohere-Embed-v4.0 & 0.277 & 0.414 & 0.329 & 0.355 & 0.324 & 0.018 & 0.009 \\
        OpenAI text-embedding-3-large & 0.110 & 0.179 & 0.111 & 0.138 & 0.111 & 0.000 & 0.000 \\
        OpenAI text-embedding-3-small & 0.116 & 0.225 & 0.087 & 0.155 & 0.087 & 0.000 & 0.000 \\
        Gemini-Embedding-2 & 0.401 & 0.567 & 0.624 & 0.560 & 0.622 & 0.019 & 0.007 \\
        \midrule
        \multicolumn{8}{l}{\textit{\textbf{Decoder-based embeddings (Qwen3)}}} \\
        Qwen3-Embedding-4B & 0.494 & 0.591 & 0.762 & 0.695 & 0.762 & 0.000 & 0.000 \\
        Qwen3-Embedding-4B-instr. & \textbf{0.543} & 0.616 & 0.836 & \textbf{0.764} & 0.836 & 0.000 & 0.000 \\
        Qwen3-Embedding-8B & 0.483 & 0.583 & 0.803 & 0.672 & 0.803 & 0.006 & 0.001 \\
        Qwen3-Embedding-8B-instr. & 0.508 & \textbf{0.635} & \textbf{0.849} & 0.712 & \textbf{0.849} & 0.003 & 0.001 \\
        \bottomrule
    \end{tabular}
\end{table*}

\subsection{E3: Per-Law Restriction and RRF (RQ1, RQ2)}
\label{subsec:res_e3}

\textbf{Whole-corpus fusion.} Fusion does not help on the whole corpus (Table~\ref{tab:fusion-whole}). None of the four RRF combinations exceeds the best dense encoder on MRR@10: the best, word BM25 with instructed 8B, reaches $0.740$ against $0.849$. Lexical-only fusion is worse than word BM25 alone ($0.258$ vs.\ $0.361$). Four-way fusion matches instructed 8B on R@20 ($0.634$ vs.\ $0.635$) but not on R@10 ($0.474$ vs.\ $0.508$) or Hit@10 ($0.950$ vs.\ $0.990$). A plausible reason is that RRF combines a strong dense ranking with a much weaker lexical one.

\begin{table}[t]
    \centering
    \caption{Whole-corpus RRF fusion ($n{=}200$) compared with its
    components. Bold: best per column.}
    \label{tab:fusion-whole}
    \scriptsize
    \setlength{\tabcolsep}{4pt}
    \begin{tabular}{lcccc}
        \toprule
        \textbf{System} & \textbf{R@10} & \textbf{R@20} & \textbf{MRR@10} & \textbf{Hit@10} \\
        \midrule
        \multicolumn{5}{l}{\textit{\textbf{Components}}} \\
        BM25 (word)                & 0.258 & 0.379 & 0.361 & 0.780 \\
        Qwen3-8B-instr.            & 0.508 & \textbf{0.635} & \textbf{0.849} & 0.990 \\
        Qwen3-4B-instr.            & \textbf{0.543} & 0.616 & 0.836 & \textbf{1.000} \\
        \midrule
        \multicolumn{5}{l}{\textit{\textbf{RRF combinations}}} \\
        word + char-4              & 0.261 & 0.428 & 0.258 & 0.780 \\
        word + 8B-instr.           & 0.436 & 0.582 & 0.740 & 0.925 \\
        word + 4B-instr.           & 0.427 & 0.563 & 0.739 & 0.920 \\
        word + char-4 + 8B + 4B    & 0.474 & 0.634 & 0.732 & 0.950 \\
        \bottomrule
    \end{tabular}
\end{table}

\textbf{Per-law restriction.} Restricting the pool to one law (Table~\ref{tab:perlaw_full}) barely changes Family (instructed 8B MRR@10 $0.849\to0.850$) but makes Evidence measurable ($0.001\to0.087$), and the two laws then behave differently. On Family, a single dense encoder is best ($0.850$) and dense--sparse fusion is worse ($0.744$). On Evidence, the 4B encoder has the highest recall (R@10 $0.300$, R@20 $0.507$), whereas fusion of the 8B encoder with word BM25 gives the best non-decomposed MRR@10 ($0.207$). That fusion's recall is close to that of the 8B encoder alone (R@10 $0.220$ vs.\ $0.196$; R@20 $0.310$ vs.\ $0.314$), so fusion mainly improves the rank of the first correct article and changes recall little. Lexical-only fusion ties word BM25 on Family MRR@10 ($0.412$) with higher R@10 ($0.463$ vs.\ $0.399$), but it is weak on Evidence ($0.018$).

Hybrid fusion is therefore not a global default: its effect depends on the law and on the metric. Because the two laws differ in pool size (252 vs.\ 129 articles), references per case, and scored cases (200 vs.\ 161), Family and Evidence scores are not a like-for-like measure of difficulty.

\begin{table}[t]
    \centering
    \caption{Per-law baselines ($n{=}200$ Family, $n{=}161$ Evidence).
    Bold: best per column.}
    \label{tab:perlaw_full}
    \scriptsize
    \setlength{\tabcolsep}{3pt}
    \begin{tabular}{lccc|ccc}
        \toprule
        & \multicolumn{3}{c|}{\textbf{Family}} & \multicolumn{3}{c}{\textbf{Evidence}} \\
        \cmidrule(lr){2-4}\cmidrule(lr){5-7}
        \textbf{System} & \textbf{R@10} & \textbf{R@20} & \textbf{MRR@10} & \textbf{R@10} & \textbf{R@20} & \textbf{MRR@10} \\
        \midrule
        \multicolumn{7}{l}{\textit{\textbf{Dense}}} \\
        Qwen3-8B-instr.     & 0.723 & \textbf{0.880} & \textbf{0.850} & 0.196 & 0.314 & 0.087 \\
        Qwen3-4B-instr.     & \textbf{0.764} & 0.864 & 0.836 & \textbf{0.300} & \textbf{0.507} & 0.130 \\
        \midrule
        \multicolumn{7}{l}{\textit{\textbf{Fusion}}} \\
        RRF, 8B + word BM25 & 0.613 & 0.783 & 0.744 & 0.220 & 0.310 & \textbf{0.207} \\
        RRF, word + char-4  & 0.463 & 0.672 & 0.412 & 0.069 & 0.204 & 0.018 \\
        \midrule
        \multicolumn{7}{l}{\textit{\textbf{Sparse}}} \\
        BM25 (word)         & 0.399 & 0.558 & 0.412 & 0.088 & 0.191 & 0.048 \\
        BM25 (char-4)       & 0.456 & 0.716 & 0.409 & 0.083 & 0.130 & 0.036 \\
        \bottomrule
    \end{tabular}
\end{table}

\subsection{E4: Decompose-and-Refine (RQ3)}
\label{subsec:res_e4}

Table~\ref{tab:dar-all} reports DaR with the four retrievers. On Evidence, DaR with char-4 BM25 is best (MRR@10 $0.696$, R@10 $0.752$), about $3.4$ and $2.5$ times the best non-decomposed values ($0.207$ and $0.300$, Table~\ref{tab:summary}). On Family, it reaches MRR@10 $0.864$ against $0.850$ for the best baseline, and its R@10 ($0.726$) is below the best baseline ($0.764$). DaR with the 4B encoder has the highest Family R@10 ($0.778$) and the highest Evidence R@20 ($0.923$).

Table~\ref{tab:dar-effect} holds the retriever fixed. DaR helps lexical retrievers on both laws and dense retrievers on Evidence. On Family, it lowers the MRR@10 of the dense encoders by approximately $0.12$ to $0.13$ while leaving their recall almost unchanged (R@10 changes by at most 0.021), so it degrades the rank of the first correct article rather than coverage. Consequently, DaR lets a lexical retriever reach the level of dense retrieval on Family without clearly exceeding it. Under DaR the same retriever (char-4 BM25) is best on both laws, so the difference in best strategy between the two laws is specific to the non-decomposed setting.

Because DaR combines LLM-generated queries, issue splitting, and a taxonomy prior, and no control generates the three channels from the undecomposed narrative, we attribute these gains to the DaR pipeline as a whole rather than to decomposition alone.

\begin{table}[t]
    \centering
    \caption{DaR systems per route (Evidence: 161 cases, Family: 200 cases).
    Bold: best per route and column.}
    \label{tab:dar-all}
    \scriptsize
    \begin{tabular}{llccc}
        \toprule
        \textbf{Route} & \textbf{System} & \textbf{R@10} & \textbf{R@20} & \textbf{MRR@10} \\
        \midrule
        \multirow{4}{*}{Evidence}
        & DaR, BM25 (char-4) & \textbf{0.752} & 0.865 & \textbf{0.696} \\
        & DaR, BM25 (word)   & 0.707 & 0.842 & 0.661 \\
        & DaR, Qwen3-4B-instr.      & 0.750 & \textbf{0.923} & 0.469 \\
        & DaR, Qwen3-8B-instr.      & 0.674 & 0.853 & 0.386 \\
        \midrule
        \multirow{4}{*}{Family}
        & DaR, BM25 (char-4) & 0.726 & \textbf{0.928} & \textbf{0.864} \\
        & DaR, BM25 (word)   & 0.749 & 0.926 & 0.848 \\
        & DaR, Qwen3-4B-instr.      & \textbf{0.778} & 0.891 & 0.720 \\
        & DaR, Qwen3-8B-instr.      & 0.702 & 0.870 & 0.716 \\
        \bottomrule
    \end{tabular}
\end{table}

\begin{table}[t]
    \centering
    \caption{Best-performing systems from each retrieval method group, selected by MRR@10 for each law.
    ($n{=}200$ Family, $n{=}161$ Evidence).}
    \label{tab:summary}
    \scriptsize
    \setlength{\tabcolsep}{4pt}
    \begin{tabular}{lcc|cc}
        \toprule
        & \multicolumn{2}{c|}{\textbf{Family}} & \multicolumn{2}{c}{\textbf{Evidence}} \\
        \cmidrule(lr){2-3}\cmidrule(lr){4-5}
        \textbf{System} & \textbf{MRR@10} & \textbf{R@10} & \textbf{MRR@10} & \textbf{R@10} \\
        \midrule
        BM25 (word)              & 0.412 & 0.399 & 0.048 & 0.088 \\
        Dense, Qwen3-8B-instr.   & 0.850 & 0.723 & 0.087 & 0.196 \\
        Dense, Qwen3-4B-instr.   & 0.836 & \textbf{0.764} & 0.130 & 0.300 \\
        RRF, 8B + word BM25      & 0.744 & 0.613 & 0.207 & 0.220 \\
        DaR (char-4 BM25)        & \textbf{0.864} & 0.726 & \textbf{0.696} & \textbf{0.752} \\
        \bottomrule
    \end{tabular}
\end{table}

\begin{table}[t]
    \centering
    \caption{Effect of DaR with the retriever held fixed. Base: raw
    \textit{Fact} query with per-law restriction; $\Delta$ = DaR $-$ Base.
    \textcolor{green!50!black}{Green}: improvement;
    \textcolor{red!75!black}{red}: decline.}
    \label{tab:dar-effect}
    \scriptsize
    \setlength{\tabcolsep}{2.5pt}
    \begin{tabular}{lccc|ccc}
        \toprule
        & \multicolumn{3}{c|}{\textbf{MRR@10}} & \multicolumn{3}{c}{\textbf{R@10}} \\
        \cmidrule(lr){2-4}\cmidrule(lr){5-7}
        \textbf{Retriever} & Base & DaR & $\Delta$ & Base & DaR & $\Delta$ \\
        \midrule
        \multicolumn{7}{l}{\textit{Family ($n{=}200$)}} \\
        BM25 (word)      & 0.412 & 0.848 & \dpos{0.436} & 0.399 & 0.749 & \dpos{0.350} \\
        BM25 (char-4)    & 0.409 & 0.864 & \dpos{0.455} & 0.456 & 0.726 & \dpos{0.270} \\
        Qwen3-4B-instr.  & 0.836 & 0.720 & \dneg{0.116} & 0.764 & 0.778 & \dpos{0.014} \\
        Qwen3-8B-instr.  & 0.850 & 0.716 & \dneg{0.134} & 0.723 & 0.702 & \dneg{0.021} \\
        \midrule
        \multicolumn{7}{l}{\textit{Evidence ($n{=}161$)}} \\
        BM25 (word)      & 0.048 & 0.661 & \dpos{0.613} & 0.088 & 0.707 & \dpos{0.619} \\
        BM25 (char-4)    & 0.036 & 0.696 & \dpos{0.660} & 0.083 & 0.752 & \dpos{0.669} \\
        Qwen3-4B-instr.  & 0.130 & 0.469 & \dpos{0.339} & 0.300 & 0.750 & \dpos{0.450} \\
        Qwen3-8B-instr.  & 0.087 & 0.386 & \dpos{0.299} & 0.196 & 0.674 & \dpos{0.478} \\
        \bottomrule
    \end{tabular}
\end{table}

\begin{figure}[ht]
    \centering
    \includegraphics[width=1\linewidth]{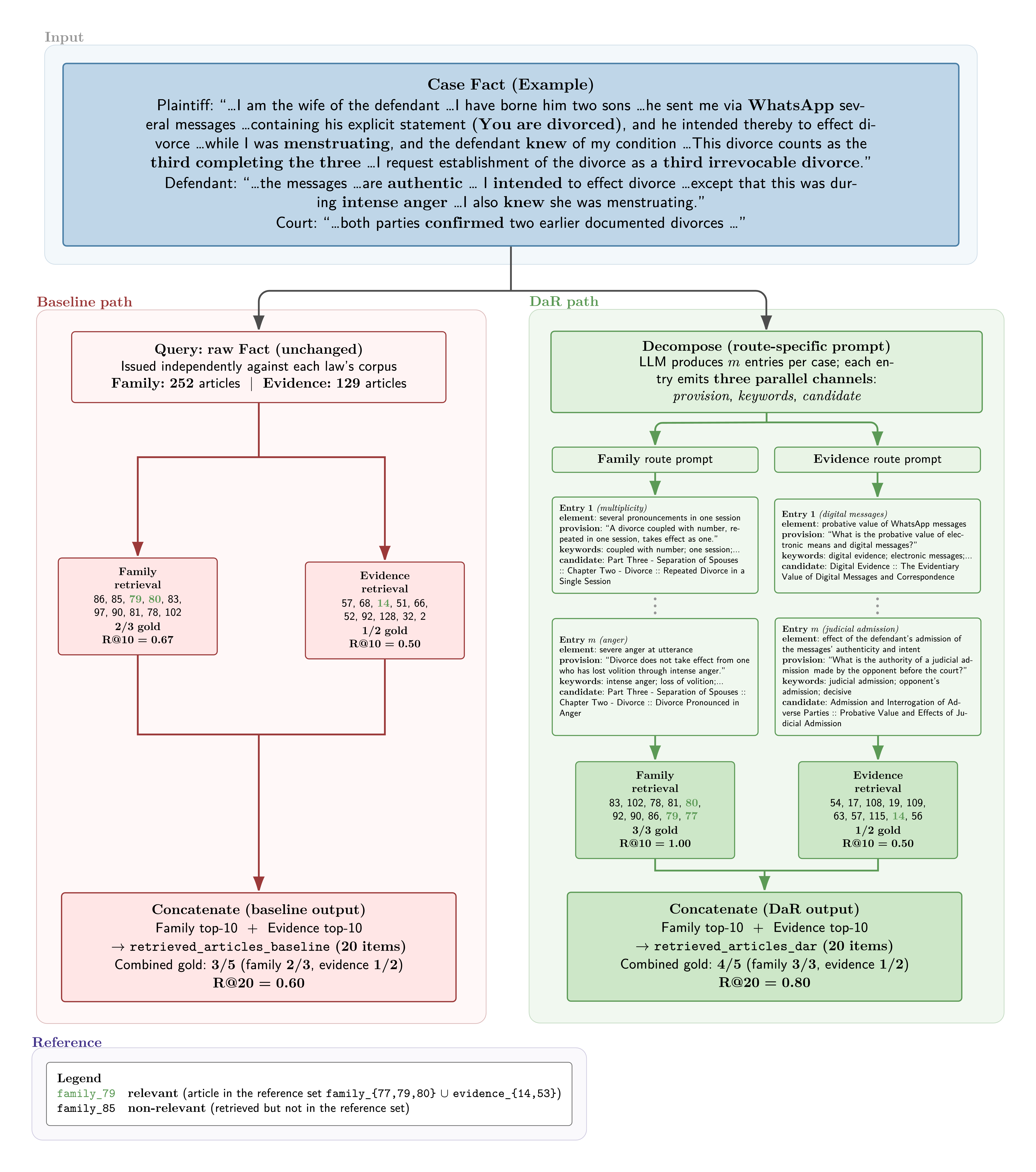}
    \caption{Baseline vs.\ DaR retrieval on one case, both routes. Green IDs are reference articles; DaR recovers one additional Family article at $K{=}10$ that the baseline misses entirely, and both paths fail to rank the case's digital-evidence article inside the top 10, though DaR alone recovers it by rank 28 of 50.}
    \label{fig:case-study}
\end{figure}

\textbf{Qualitative case study.} Fig.~\ref{fig:case-study} traces one case (five reference articles: \texttt{family\_77}, \texttt{family\_79}, \texttt{family\_80}, \texttt{evidence\_14}, \texttt{evidence\_53}) through both the baseline and DaR paths end to end. On the Family side, DaR recovers all three reference articles at $K{=}10$ (R@10 $=1.00$) against the baseline's two of three (R@10 $= 0.67$), correctly surfacing \texttt{family\_77} -- the general definition of divorce -- which the baseline never ranks in its top 10. On the Evidence side, both paths retrieve \texttt{evidence\_14} (judicial admission) but neither ranks \texttt{evidence\_53} (digital-evidence admissibility, the article governing the WhatsApp messages at the center of the case) inside the top 10. Extending to $K{=}50$ makes the two paths' failure qualitatively different rather than merely a matter of degree: \texttt{evidence\_53} never appears in the baseline's top 50 at all, while DaR surfaces it at rank 28. This is compatible with the generated channels moving the query toward the statute's vocabulary. Concatenating both laws' top-10 lists (the corpus's combined export format) raises Recall@20 from $0.60$ (baseline) to $0.80$ (DaR) on this case -- consistent with the aggregate DaR gains in Table~\ref{tab:dar-all}, though a single case cannot establish significance on its own.

\textbf{Limitations.} The 200 narratives are synthetic and may not capture the variability of real judicial records, and citation-derived labels may omit applicable but uncited provisions. The corpus covers two Saudi laws and divorce-establishment scenarios only. Evidence is scored only on the 161 cases that cite it, so the cost of invoking the Evidence route on the other 39 cases is not measured. Decomposition uses one LLM run per case (\texttt{google/gemini-3.6-flash}~\cite{google2026gemini36flash}, temperature $0$), so run-to-run variance is unknown. The provision channel may draw on the model's memory of public statutes, the candidate channel may exploit predictable chapter names. Because DaR bundles LLM query generation with decomposition, its gains cannot be attributed to decomposition alone, and the three channels were not evaluated separately; a channel-level ablation and an undecomposed control are needed to isolate their respective contributions. The findings may not generalize to other legal domains.

\section{Conclusion}
\label{sec:conclusion}

We evaluated sparse, dense, hybrid, and decomposition-based retrieval of statutory articles from Saudi Arabic judicial narratives. The strongest dense encoder owes its overall result to the Family Law: on the Law of Evidence alone it falls sharply, and fusion with BM25 gives the best non-decomposed Evidence result but remains weak. DaR substantially improves Evidence retrieval and yields a small increase in the best Family MRR@10. Without decomposition, the two laws favor different strategies; under DaR, the same retriever is best on both. Limitations include the synthetic nature of the benchmark, citation-derived labels, coverage of only two laws and divorce-establishment scenarios, unmeasured Evidence-route cost on non-Evidence cases, and single-run decomposition. Future work will expand the benchmark with more cases and additional Saudi laws, and will repeat decomposition runs to assess variance.

\section*{Conflict of Interest}
The authors declare that there is no conflict of interest regarding the publication of this paper.

\bibliographystyle{IEEEtran}
\bibliography{sample.bib}
\end{document}